\documentclass[lettersize,journal]{IEEEtran}
\usepackage{amsmath,amsfonts}
\usepackage{algorithmic}
\usepackage{algorithm}
\usepackage{array}
\usepackage[caption=false,font=normalsize,labelfont=sf,textfont=sf]{subfig}
\usepackage{textcomp}
\usepackage{stfloats}
\usepackage{url}
\usepackage{verbatim}
\usepackage{graphicx}
\usepackage{cite}
\usepackage{array}
\usepackage{doi}
\usepackage{color}
\usepackage{amssymb}
\usepackage{float}
\usepackage{amsthm,amsmath,amssymb}
\usepackage{mathrsfs}
\usepackage{hyperref}

\usepackage{algorithm}
\usepackage{algorithmic}
\usepackage{multirow}
\usepackage{amsmath}
\usepackage{graphicx} 
\usepackage{color}
\usepackage[subfigure]{tocloft}
\begin{document}

\title{FedLED: Label-Free Equipment Fault Diagnosis with Vertical Federated Transfer Learning}

\author{Jie Shen, Shusen Yang, Cong Zhao, Xuebin Ren, Peng Zhao, Yuqian Yang, Qing Han, and Shuaijun Wu 
}

\IEEEpubid{}

\maketitle

\begin{abstract}
Intelligent equipment fault diagnosis based on Federated Transfer Learning (FTL) attracts considerable attention from both academia and industry.
It allows real-world industrial agents with limited samples to construct a fault diagnosis model without jeopardizing their raw data privacy.
Existing approaches, however, can neither address the intense sample heterogeneity caused by different working conditions of practical agents, nor the extreme fault label scarcity, even zero, of newly deployed equipment.
To address these issues, we present FedLED, the first unsupervised vertical FTL equipment fault diagnosis method, where knowledge of the unlabeled target domain is further exploited for effective unsupervised model transfer.
Results of extensive experiments using data of real equipment monitoring demonstrate that FedLED obviously outperforms SOTA approaches in terms of both diagnosis accuracy (up to 4.13$\times$) and generality.
We expect our work to inspire further study on label-free equipment fault diagnosis systematically enhanced by target domain knowledge.
\end{abstract}


\section{Introduction}
Proliferating data-driven methods have been proven to be promising in intelligent equipment fault diagnosis~\cite{sun2017intelligent},
where the diagnosing process is usually modeled as classifying `normal' and `fault' samples with multiple features extracted from various equipment monitoring signals like current, voltage, vibration, temperature, and acoustic emission~\cite{islam2019reliable}.
Predominating approaches rely on abundant well-labeled samples to train various Machine-Learning (ML) models (\emph{e.g.}, SVM, DNN) that significantly outperform conventional diagnosis methods based on partial system mechanisms or expert experiences.

However, the application of such labeled sample-intensive methods is severely restricted by the extreme scarcity of fault samples in a wide range of industrial equipment holders (referred to as agents below) in practice \cite{zhao2021signal,lou2022machinery,zhao2021applications}. 
For example, for any smart manufacturer with a piece of newly deployed equipment, considering that equipment faults are generally small probability events, it usually takes a considerably long time before a single fault occurs and a sample with a determined `fault' label can be collected \cite{lei2018machinery}.
It becomes a common bottleneck for agents to construct an effective fault diagnosis model with few or even zero fault label.

To address this issue, an intuitive idea is to train a model at other agents possessing the same type of equipment with more well-labeled samples (\emph{i.e.}, the source agent), then deploy the trained model at the sample-scarce agent (\emph{i.e.}, the target agent), 
under the assumption that samples from different agents are Independently and Identically Distributed (IID).
Unfortunately, such an assumption is almost impossible to be guaranteed in practice~\cite{xia2022moment}.
Due to multi-folded differences such as monitoring setup, working load, transmission path, noise interference, and fault degree, samples from agents with different working conditions inevitably have obvious distribution discrepancies~\cite{qian2023deep}, \emph{i.e.}, different agents usually have different \textit{sample domains}.
Transfer Learning~\cite{liang2019intelligent} is primarily used to construct and transfer models across different domains, which has been applied to equipment fault diagnoses~\cite{chen2023deep}.
However, existing approaches require exchanging samples between the source and target agents for common knowledge extraction, which poses serious threats to agents' data privacy~\cite{ma2021asynchronous} considering harsh regulations like the General Data Protection Regulation (GDPR)~\cite{regulation2018general}.
Therefore, Federated Transfer Learning (FTL) methods~\cite{chen2022federated} emerge and enable transfer learning across `data islands' of different agents, where all raw samples remain local and only intermediate learning results (\emph{e.g.}, gradients) are exchanged between agents.

As illustrated in Fig.\ref{fig:setting}, to construct a fault diagnosis model under the aforementioned practical scenario, existing FTL methods, however, cannot be directly applied due to the following two fundamental issues of real-world equipment monitoring data:

\begin{figure}[t]
    \centering
    \includegraphics[width=3in]{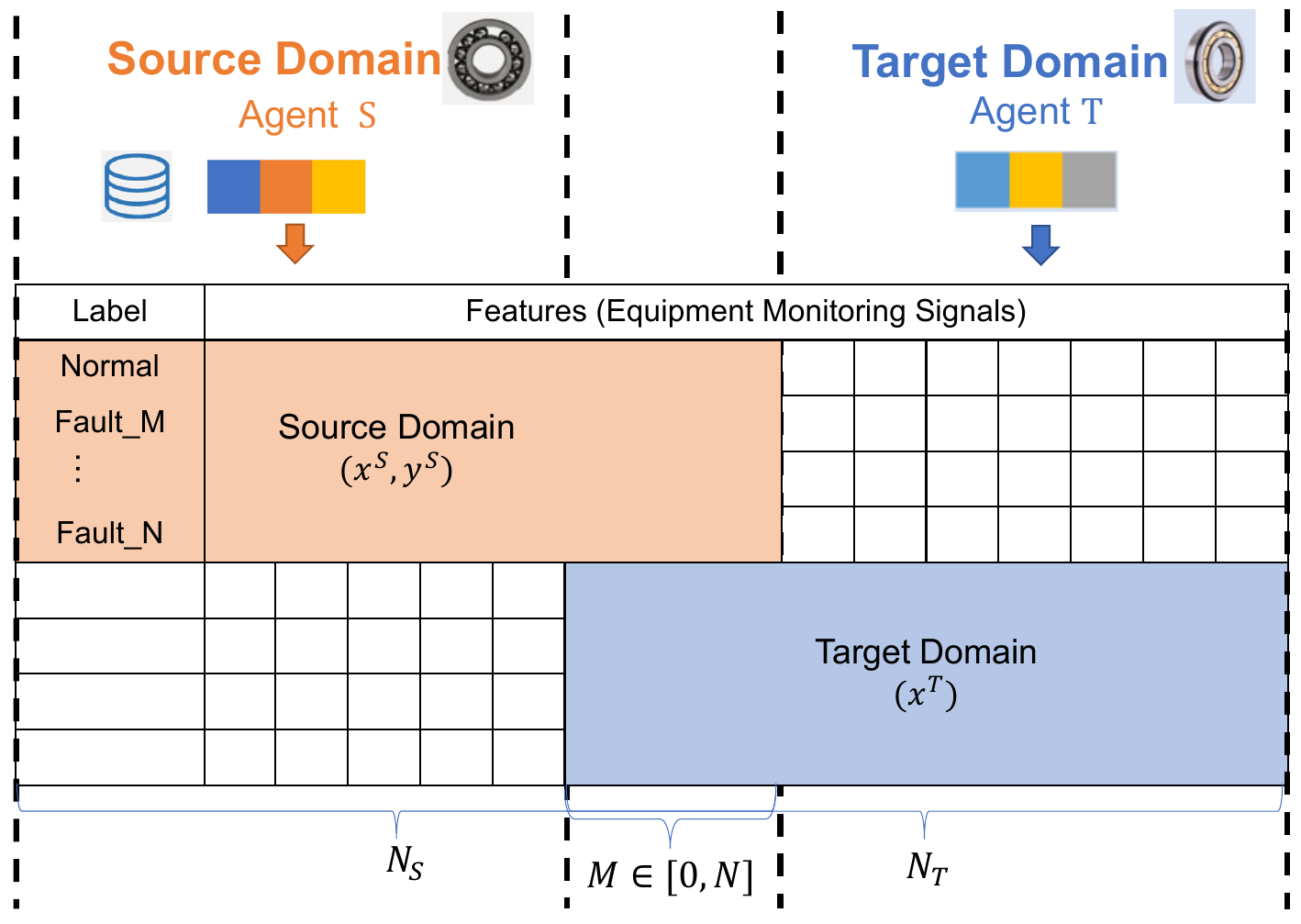}
    \caption{The General Equipment Fault Diagnosis Scenario based on Vertical Federated Transfer Learning}
    \label{fig:setting}
\end{figure}

\begin{enumerate}
\item \textbf{Intense Sample Heterogeneity:}
ML-based FTL methods like~\cite{yang2021federated} require overlapping samples (\emph{i.e.}, the same set of samples possessed by both the source and target agents) to achieve effective model transfer.
This is almost impossible in practice since the two agents cannot possess overlapping samples (from the same piece of equipment under the same monitoring setup) without exchanging raw data.
Fault diagnosis with Deep Learning (DL)-based FTL~\cite{zhang2022data} requires no overlapping samples, but
assumes that both agents share the same feature space, \emph{i.e.}, horizontal FTL, where both agents use the same set of sensors with the same monitoring deployment.
However, in practice, feature spaces of different agents are often multi-scale~\cite{yang2022massive} and intensely heterogeneous~\cite{islam2019reliable} due to the agents' different monitoring setups.
For example, the source samples $x^S$ are collected by $N_S$ sensors while the target samples $x^T$ are
collected by $N_T$ sensors, where only $M$ ($0 \leq M \leq \min{(N_S, N_T)}$) sensors are shared.
Fault diagnosis based on vertical FTL is of great significance.

\item \textbf{Zero Fault Label:}
Most FTL methods like~\cite{liu2020secure} require a small set of labeled samples in the target domain for model training.
However, it is highly likely for practical agents, especially those with newly deployed equipment, to have zero fault label~\cite{yang2022multi,chai2021multisource,zhao2021applications} (\emph{i.e.}, with no $y^T$).
The only unsupervised FTL fault diagnosis approach~\cite{tzinis2021separate} requiring no target domain label is restricted by the same feature space assumption and cannot be applied to the aforementioned vertical FTL scenario.
\end{enumerate}

To address the above issues, we present the first unsupervised vertical FTL equipment fault diagnosis method FedLED.
It enables model transferring to the target domain with zero fault label from an intensely heterogeneous source domain.
The main contributions of this paper are as follows:

\begin{enumerate}
\item
We are the first to concentrate on the problem of transferring the equipment fault diagnosis model between agents with heterogeneous feature spaces and zero target domain fault label, which is a common bottleneck for the industry.
A new fault diagnosis method FedLED based on unsupervised vertical FTL is proposed, which can serve a wide range of agents, especially those with newly deployed equipment.

\item
In FedLED, a vertical federated joint domain adversarial adaptation  is proposed to map heterogeneous source and target features to a public latent feature space.
To enhance the effectiveness of zero fault label model transferring, we construct a novel joint domain alignment that minimizes the distance between the source label distribution and the target classification result distribution, fundamentally different from the conventional pseudo label method that does not comprehensively leverage the target domain information.

\item
We conducted extensive experiments using fault datasets of different real equipment (\emph{i.e.}, two bears and a gear) to comprehensively validate the effectiveness of our approach.
Experimental results demonstrate that FedLED prominently outperforms state-of-the-art methods in diagnosis accuracy (up to $4.13\times$ higher) under various vertical FTL scenarios.
Furthermore, FedLED stably maintains the highest diagnosis accuracy among all comparatives under different vertical FTL settings (\emph{i.e.}, sample/feature overlapping ratios), and shows more obvious advantages under harsher ones (\emph{e.g.}, when there is zero sample/feature overlapping).
\end{enumerate}

The rest of the paper is organized as follows.
Section 2 discusses related work of intelligent equipment fault diagnosis.
The system model and problem definition are provided in Section 3.
Section 4 presents FedLED in detail.
Experimental results are provided and comprehensively discussed in Section 5.
We conclude the paper in Section 6.

\section{Related Work}
In this section, we discuss existing efforts on intelligent equipment fault diagnosis.

\subsection{Intelligent Equipment Fault Diagnosis}
In recent years, intelligent data-driven equipment fault diagnosis has largely benefited from the successful development of deep learning, which has been attracting the industry due to its high diagnostic accuracy \cite{li2020industrial}. 
However, existing methods usually rely on abundant well-labeled data or IID assumption \cite{han2019novel,lu2017fault}, severely limiting their usability in practice.
Transfer learning can be used to assist training with Non-IID samples from other related agents under different scenarios.

\subsection{Transfer Learning for Equipment Fault Diagnosis}
\subsubsection{Heterogeneous Transfer Learning}
Most existing studies follow the same distribution assumption that is difficult to satisfy in practice \cite{rauber2014heterogeneous}.
Heterogeneous sample processing methods in the field of fault diagnosis mainly process heterogeneous features through feature screening and other methods, and then substitute them into traditional machine learning methods (such as SVM \cite{van2021data}, KNN, \emph{etc.}) for training.

\subsubsection{Unsupervised Transfer Learning}
Widely adopted unsupervised transfer learning methods can be divided into discrepancy-based and adversarial-based unsupervised transfer learning.
Discrepancy-based methods  align the source and target domains by measuring the data distributions distance. Common methods for distribution distance measuring  include correlation alignment (CORAL) \cite{sun2016deep}, maximum mean discrepancy (MMD)  \cite{sejdinovic2013equivalence}, and joint distribution adaptation (JMMD) \cite{long2013transfer}.
Adversarial-based methods use a domain discriminator to reduce the feature distribution discrepancy between source and target domains produced by the feature extractors, enabling cross-machine troubleshooting.
Predominating methods include domain adversarial neural network (DANN) \cite{ganin2016domain} and conditional domain adversarial network (CDAN) \cite{long2018conditional}. 

Most of the transfer learning methods need to obtain shared knowledge by accessing the raw data of the source  and the target domains, casting serious threats to data privacy.
FTL methods emerge \cite{yang2019federated} to address the data privacy issue.

\subsection{Federated Transfer Learning for Equipment Fault Diagnosis}
Traditional FTL methods \cite{yang2019federated} require shared samples  and a small number of labels in the target domain, which limited their use in practical scenarios.
With the development of AI technology, deep learning methods have gradually become mainstream.
\cite{peng2019federated} proposes to address domain drift in federated learning based on adversarial domain adaptation.
\cite{chen2020fedhealth} provides an FTL system that utilizes prior distributional knowledge to reduce inter-domain gaps.

However, all studies above concentrate on horizontal federated learning, where the source and target domains share the same feature space.
Their performance cannot be guaranteed in the vertical FTL scenario.

\section{System Model and Problem Definition}
In this section, we provide the general system model of fault diagnosis based on vertical FTL in Fig.\ref{fig:setting}, and the formal definition of our research problem.

\subsubsection{System Model}

Taking the scenario in Fig.\ref{fig:setting} as an example, there are two vertical FTL agents: the source domain agent S and the target domain agent T.
For the well-labeled source domain, $D^S=\{(x_{i}^S,y_{i}^S)\}_{i=1}^{N^S}$, where $N^S$ is the total amount of samples of $D^S$.
$x_{i}^S$ and $y_{i}^S$ denote the $i$th sample in $D^S$ and its label, respectively.
$x_{i}^S \in \mathbb{R}^{N_{S}}$, where $\mathbb{R}^{N_{S}}$ is the source feature space, and $N_{S}$ represents the feature number.
For the label-free target domain, $D^T=\{x_i^T\}_{i=1}^{N^T}$, $x_i^T \in \mathbb{R}^{N_{T}}$, where $N^T$ is the sample number, $\mathbb{R}^{N_{T}}$ represents the target feature space, and $N_{T}$ is the feature number.
Considering that both the source and target domain agents focus on the same types of faults of the same type of equipment, we assume that both the source and target domains follow the same fault distribution in the same label space.

Considering the heterogeneous source and target domains in practice, there are 1) $D^S \cap D^T = \emptyset$, \textit{i.e.}, no overlapping samples, and 2) $M \in [0, N]$, where $M$ denotes the number of overlapping features between $\mathbb{R}^{N_{S}}$ and $\mathbb{R}^{N_{T}}$, and $N=\min{(N_{S}, N_{T})}$.
It is highly likely that such feature space heterogeneity induces a non-negligible distance $dist(\mathbb{R}^{N_{S}}, \mathbb{R}^{N_{T}}) > \epsilon > 0$.

\subsubsection{Problem Definition}
The vertical FTL task is a classification problem in machine learning with classifier $\mathcal{F_C}$. 
Considering our system model, there are two constraint conditions must be met:
1) intense sample heterogeneity, $D^S \cap D^T = \emptyset$ and $dist(\mathbb{R}^{N_{S}}, \mathbb{R}^{N_{T}}) > \epsilon$, and 
2) zero fault label, $\{y^T\}$ is unavailable. 
The vertical FTL task can be defined as a constrained optimization problem:
\begin{equation}
\begin{aligned}
\min_{\mathcal{F_C}}{}&Loss(\mathcal{F_C}(x), y)\\
s.t. &\left \{
\begin{array}{l}
    D^S \cap D^T = \emptyset, \\
    dist(\mathbb{R}^{N_{S}}, \mathbb{R}^{N_{T}}) > \epsilon,\\
    \{y^T\}= \emptyset.           
\end{array}
\right.     
\end{aligned}
\label{con:ori}
\end{equation}

Considering the intense sample heterogeneity, the feature-based domain adaptation method can be used to build feature extractors $\mathcal{F_S}$ and $\mathcal{F_T}$ to map the source and target domains to a latent common space.
For zero fault label, the source and target domain output label distributions $(\mathcal{F_C} (\mathcal{F_S}), \mathcal{F_C} (\mathcal{F_T}))$ can be aligned, since their label spaces are assumed to be the same. 
Our research problem can be  transformed into Eq.(\ref{con:base_question1}):
\begin{equation}
\begin{aligned}
    \min_{\mathcal{F_C}, \mathcal{F_S}, \mathcal{F_T}} &Loss(\mathcal{F_C} (\mathcal{F_S}(x^S)), y^S) + \lambda dist(\mathcal{F_S}(x^S), \mathcal{F_T}(x^T)) \\
   &+ \beta  dist(\mathcal{F_C} (\mathcal{F_S}(x^S)), \mathcal{F_C} (\mathcal{F_T}(x^T))).
   \label{con:base_question1}
\end{aligned}
\end{equation}
The supplementary details of problem definition is presented in Supplemental Material \footnote{[Online Available]: \href{https://github.com/htkg987/FedLED/blob/main/supplemental_material.pdf}{https://github.com/htkg987/FedLED/blob/main/supple-\\
mental\_material.pdf}} S.1.

\section{Label-Free Equipment Fault Diagnosis with Vertical FTL}
In this section, considering our system model and problem definition, we present an unsupervised vertical FTL equipment fault diagnosis method FedLED.
It comprises a vertical federated transfer model that can train a target domain fault diagnoser without fault labels,
and an unsupervised federated model training scheme.

\begin{figure}[t]
    \centering
    \includegraphics[width=3.8in]{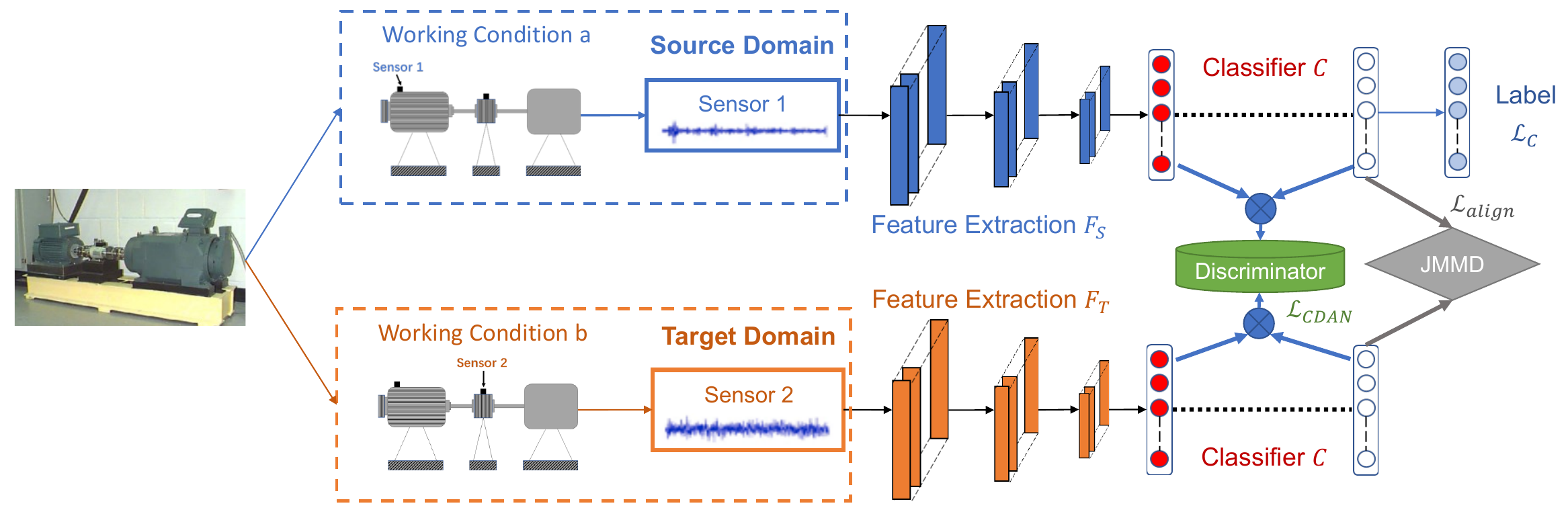}
    \caption{The Structure of Unsupervised Vertical Federated Transfer Model}
    \label{fig:model1}
\end{figure}

\subsection{Model Structure}
The problem in Eq.(\ref{con:base_question1}) can be modeled as an unsupervised domain adaptation problem, and there are three joint-optimization objectives, \textit{i.e.}, the classification loss, the feature alignment loss, and the output label alignment loss.
The overall network structure is shown in  Fig.\ref{fig:model1}. Considering the two constraint conditions of vertical FTL, our unsupervised vertical federated transfer model structure consists of two parts:
1) the vertical federated joint domain adversarial adaptation for the calculation of the classification and feature alignment losses, and 
2) the joint domain alignment to calculate the output label alignment loss.

We propose a vertical federated joint domain adversarial adaptation, based on the Adversarial-based method CDAN and vertical federated scheme. The key is a novel conditional domain discriminator conditioned on the cross-covariance of domain-specific feature representations and classifier predictions, which can map heterogeneous source and target feature spaces to a latent common space.
Even if the discriminator is completely obfuscated, there is no guarantee that the feature extractor can extract domain-invariant features.  Since the domain adversarial adaptation has already aligned the feature space, we additionally add the  discrepancy-based joint alignment method as the  joint domain alignment to calculate the output label alignment loss, which minimizes the distance between the source label distribution and the target classification result distribution, fundamentally different from the conventional pseudo label method that does not comprehensively leverage the target domain information. 
More supplementary details of model construction are shown in Supplemental Material S.2.

\subsubsection{Vertical Federated Joint Domain Adversarial adaptation for Intense Sample Heterogeneity}
The vertical federated joint domain adversarial adaptation is based on adversarial-based CDAN \cite{arora2017generalization} to calculate the classification loss and feature alignment loss. The key to CDAN  is a novel conditional domain discriminator conditioned on the cross-covariance of feature representations and classifier predictions, which can extract domain-invariant features from heterogeneous features.

\textbf{Feature alignment loss:} It is first defined as a minimax optimization problem with two competing error terms, and the overall objective function is as follows:
\begin{equation}
\begin{aligned}
    &\mathcal{L}_{CDAN}(\theta_{F_S}, \theta_{F_T}, \theta_D, \theta_C) \\
    =& \mathbb{E}_{x^S_i\sim D^S} W(P^S_i)\log{[D(f_S\otimes g_S)]}\\
    &+\mathbb{E}_{x^T_j\sim D^T} W(P^S_j)\log{[1-D(f_T\otimes g_T)]},
    \label{con:eqjdan}
\end{aligned}
\end{equation}
where $f_S$ and $g_S$ represent the high-level features of the source domain and the output of the classifier through the high-level features, respectively, and $f_T$, $g_T$ correspond to the high-level features of the target domain and their outputs on the classifier.  $\otimes$ represents a multi-linear map, which represents the outer product of multiple random vectors. The joint distribution $P(x, y)$ of any two random vectors $x$, $y$ can be obtained by using the cross covariance $( \mathbb{E}_{xy}[\Phi(x) \otimes \Phi(y)])$, where $\Phi$ represents the reproducible kernel function. At the same time, an additional dynamic sample weight method is used to avoid negative samples from affecting training. The update method of the sample weight is as Eq.(\ref{con:eqwei}), where $p$ represents the probability that the classifier finally predicts each category:
\begin{equation}
    W(p)=1+e^{\sum_{c=0}^{N_C-1}}{p_c\log{y_c}}.
    \label{con:eqwei}
\end{equation}

The optimization method of joint domain adversarial learning is: by minimizing (\ref{con:eqjdan}), optimize the parameters of classifier $C$ and feature extractor $F$ ($F_S$, $F_T$), while maximizing (\ref{con:eqjdan}) to optimize the domain discriminator $D$, the objective function of each model as follows:
\begin{equation}
    (\hat{\theta^{t+1}_F}, \hat{\theta^{t+1}_C})=arg\min_{\theta_F,\theta_C}\mathcal{L}(\theta^t_F,\theta^t_C,\hat{\theta^t_d}),
\end{equation}
\begin{equation}
    \hat{\theta^{t+1}_D}=arg\max_{\theta_D}\mathcal{L}(\hat{\theta^t_F},\hat{\theta^t_C},\theta^t_d).
\end{equation}

\textbf{Classification loss:} In order to avoid the task difference between the target domain and the source domain on the domain-invariant features, the final global classification task is weakened. Therefore, the supervised learning method on source domain is added to prevent the classification bias of the classifier. The objective function of the supervised classification task is as follows:
\begin{equation}
    \mathcal{L}_C(\theta_{F_S},\theta_C)= \mathbb{E}_{(x_i^S, y_i^S) \in D^S}\sum^{N^S-1}_{i=0}\mathcal{L}(C(F_S(x_i^S)),y_i^S). 
    \label{con:eqC}
\end{equation}

\subsubsection{Joint Domain Alignment for Zero Fault Label}
Recent work \cite{arora2017generalization} reveals that even if the discriminator is completely obfuscated, there is no guarantee the feature extractor can extract domain-invariant features. This risk arises from the equilibrium challenges that exist in adversarial learning. Since CDAN has already aligned the feature space, we also add the discrepancy-based alignment \cite{long2017deep} method as the joint domain alignment to calculate the output label alignment loss. 

\textbf{Output label alignment loss:} The objective function of this joint domain align process is as follows:
\begin{equation}
    \mathcal{L}_{align}=\| \mathbb{E}_{f_S} [\otimes^{|L|}_{l=1}\phi^{l}(g^S_{l})]-\mathbb{E}_{f_T} [\otimes^{|L|}_{l=1}\phi^{l}(g^T_{l})] \|^2_{\otimes^{|L|}_{l=1}\mathcal{H}^l}.
    \label{con:eqalign}
\end{equation}
where $g_l^S$ represents the input of high-level features in the source domain into the classifier network, its output on the $l$-th layer, and $\otimes^{|L|}_{l=1}\phi^{l}(g_{l})=\phi^{1}(g_{1} \otimes , \dots, \phi^{L}(g_{L}))$ indicates that the output of each layer of the classifier is projected into a Hilbert space through multidimensional linear mapping. $|L|$ indicates the number of layers in the classifier, usually chooses the last two layers of classifier output for fault diagnosis task alignment on different domains.

\subsection{The Unsupervised Federated Training Scheme}
The entire federated training scheme is divided into two steps in the training process: federated model initialization and federated model training.

\subsubsection{Federated Model Initialization}
The federated initialization adopts the pre-training-fine-tuning method demonstrated as effective in \cite{hendrycks2019using}.
Compared with training from scratch, pre-training the model reduces training time and speeds up the training convergence.
The result of pre-training is only a preliminary improvement to prevent overfitting.
The objective function of the federated initialization phase is defined as follows:
\begin{equation}
    \mathcal{L}_{pre} = \mathbb{E}_{(x^S_i, y^S_i)\sim D^S}\mathcal {L}(C(F_S(x^S_i; \theta_{f_S});\theta_C),y^S_i).
    \label{con:eqc}
\end{equation}
Our federated initialization process is shown in Algorithm \ref{con:alg_initial}, where the labeled source domain and label-free target domain are initialized with pre-training and randomly, respectively.

\begin{algorithm}[t]  
\small{
\caption{ Federated Model Initialization}  
\textbf{Input:} Source data $D^S$, randomly initialized source domain model $\theta_{F_S}$, $\theta_C$,  learning rate $\mu$.\\  
\textbf{Output:} Local model parameters $(\theta_{F_S},\theta_C)$ with completed federated initialization \\  
\textbf{For} epoch = 1 to $N_{epoch}^{pre}$
\textbf{do}  
\begin{enumerate}   
\item Randomly select a portion of the source domain data from $D^S$.
\item Forward propagation calculation (\ref{con:eqc}).
\item Backpropagation updates the source domain target extractor, $\theta^{t+1}_{F_S}=\theta^{t}_{F_S}-\mu (\frac{\partial \mathcal{L}_{pre}}{\partial \theta^{t}_{F_S}})$.
\item Backpropagation updates the source domain classifier, $\theta^{t+1}_{C}=\theta^{t}_{C}-\mu (\frac{\partial \mathcal{L}_{pre}}{\partial \theta^{t}_{C}})$.
\end{enumerate}   
\textbf{End For}
\label{con:alg_initial}  }
\end{algorithm}

\begin{figure}[!t]
    \centering
    \includegraphics[width=3.5in]{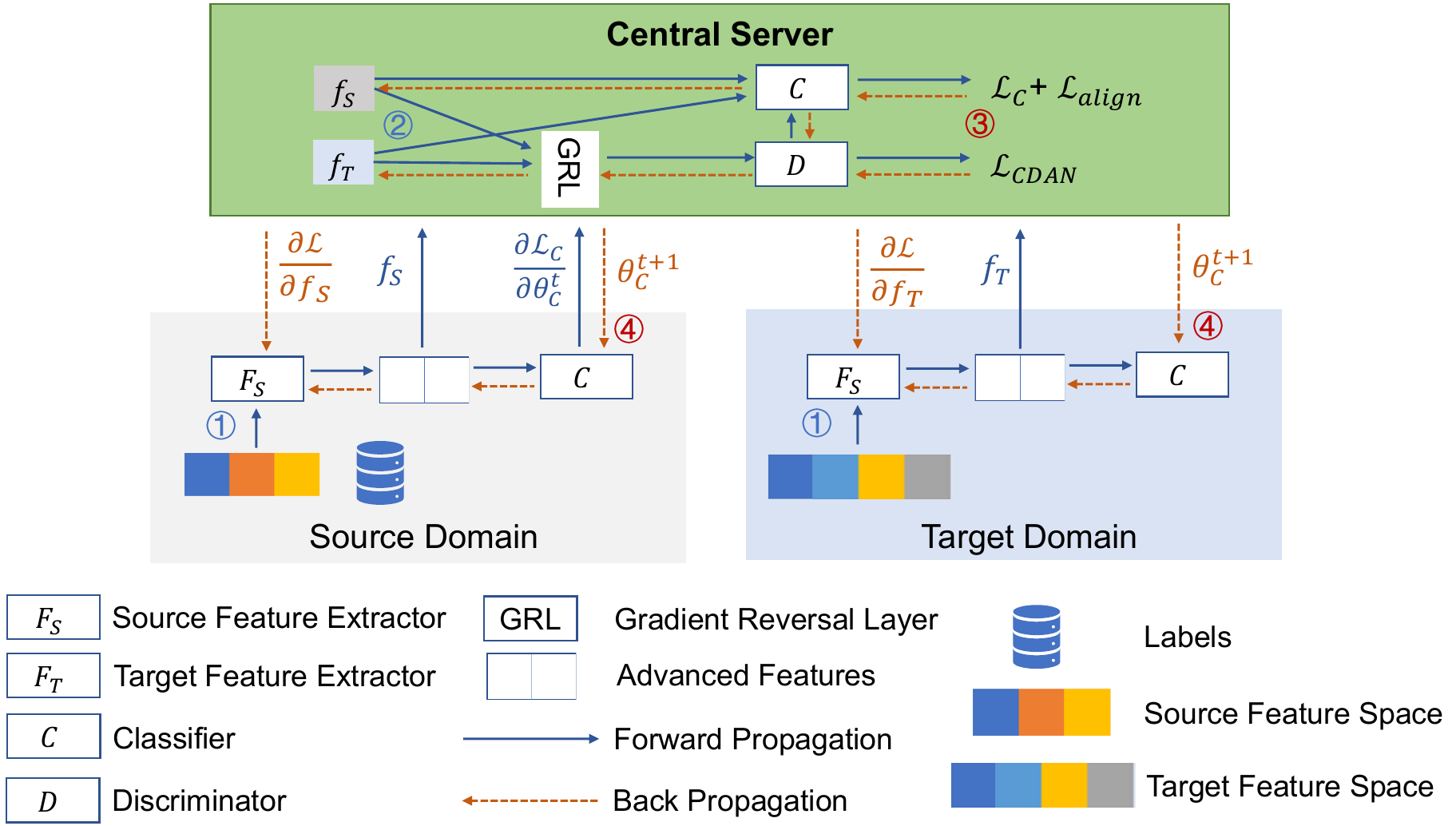}
    \caption{The Overall Workflow of Federated Model Training}
    \label{fig:liuchen}
\end{figure}

\subsubsection{Federated Model Training}
In the federated model raining process, the central server calculates the corresponding loss and gradient, then transmits the gradient to the corresponding participants.
The overall workflow is shown in Fig.\ref{fig:liuchen}.

Considering the model structure, the objective function of FedLED training is shown in Eq.(\ref{con:eqall}).
\begin{equation}
\begin{aligned}
    &\mathcal{L}_C(\theta_{F_S},\theta_{F_T},\theta_C,\theta_D)\\
    = &\mathcal{L}_C(\theta_{F_S},\theta_C) - 
    \lambda \mathcal{L}_{CDAN}(\theta_{F_S}, \theta_{F_T}, \theta_D, \theta_C) \\
    &+  \beta \mathcal{L}_{align}(\theta_{F_S}, \theta_{F_T}, \theta_C),
    \label{con:eqall}
\end{aligned}
\end{equation}
where $\lambda$ and $\beta$ are two hyperparameters, network parameters are updated during training using the Adam optimizer, and the adversarial network optimization problem is solved using a gradient inversion layer \cite{li2019diagnosing}. During each training iteration, parameters are updated as follows:
\begin{equation}
 \begin{aligned}
    \theta^{t+1}_{F_S}= &\theta^{t}_{F_S}-\mu(\frac{\partial \mathcal{L}_C }{\partial \theta^{t}_{F_S}} - \lambda \frac{\partial \mathcal{L}_{CDAN} }{\partial \theta^{t}_{F_S}} + \beta \frac{\partial \mathcal{L}_{align} }{\partial \theta^{t}_{F_S}})\\
    = &\theta^{t}_{F_S}-\mu(\frac{\partial \mathcal{L}_C }{\partial f_S} \frac{\partial f_S}{\partial \theta^{t}_{F_S}} \\
    &- \lambda \frac{\partial \mathcal{L}_{CDAN} }{\partial f_S} \frac{\partial f_S}{\partial \theta^{t}_{F_S}} + \beta \frac{\partial \mathcal{L}_{align} }{\partial f_S} \frac{\partial f_S}{\partial \theta^{t}_{F_S}}),
    \label{con:eqfs}
\end{aligned}
\end{equation}
\begin{equation}
    \theta^{t+1}_{F_T}=
    \theta^{t}_{F_T}-\mu( - \lambda \frac{\partial \mathcal{L}_{CDAN} }{\partial f_T} \frac{\partial f_T}{\partial \theta^{t}_{F_T}} + \beta \frac{\partial \mathcal{L}_{align} }{\partial f_T} \frac{\partial f_T}{\partial \theta^{t}_{F_T}}),
    \label{con:eqft}
\end{equation}
\begin{equation}
    \theta^{t+1}_C=\theta^{t}_C-\mu(\frac{\partial \mathcal{L}_C }{\partial \theta^{t}_C} - \lambda \frac{\partial \mathcal{L}_{CDAN} }{\partial \theta^{t}_C} + \beta \frac{\partial \mathcal{L}_{align} }{\partial \theta^{t}_C}),
    \label{con:eqcc}
\end{equation}
\begin{equation}
    \theta^{t+1}_D=\theta^{t}_D-\mu(- \lambda \frac{\partial \mathcal{L}_{CDAN} }{\partial \theta^{t}_D} ).
    \label{con:eqd}
\end{equation}
Here, $\mu$ is the learning rate and $t$ represents the $t$-th iteration update.
Our training process is described in Algorithm \ref{con:alg_tra}.

\begin{algorithm}[t]  
\small{
\caption{\small Federated Model Training}  
\textbf{Input:} Source data $D^S$, target data $D^T$, learning rate $\eta$,  hyperparameters $(\lambda,\beta)$,  local model of central server  $(\theta_D, \theta_C)$, local model of source domain $(\theta_{F_S}, \theta_C)$, local model of target domain $(\theta_{F_T}, \theta_C)$. \\  
\textbf{Output:} The trained local model of target domain $(\theta_{F_T}, \theta_C)$ . \\  
\textbf{For} epoch = 1 to $N_{epoch}^{train}$
\textbf{do}  \\
\textbf{Source Domain} 
\begin{enumerate} 
\item Randomly select  source domain sample $X_S$ from $D^S$.
\item Obtain the advanced feature $f_S=F_S(X^S)$ and classification loss, send the advanced feature and gradient information $\frac{\partial \mathcal{L}_C}{\theta^{t}_C}$ to the central server.
\item Block waiting for $\frac{\partial \mathcal{L}}{\partial f_S}$ and $\theta^{t+1}_C$  from the central server.
\item Update feature extractor according to Eq.(\ref{con:eqfs}), and overwrite the current classifier parameters with the $\theta^{t+1}_C$ .
\end{enumerate}  
\textbf{Target Domain}
\begin{enumerate} 
\item Randomly select  target domain sample $X_T$ from $D^T$.
\item Obtain the advanced feature $f_T=F_T(X^T)$ corresponding to $X^T$, and send the advanced feature to the central server.
\item Block waiting for $\frac{\partial \mathcal{L}}{\partial f_T}$ and $\theta^{t+1}_C$  from the central server.
\item Update feature extractor according to Eq.(\ref{con:eqft}), and overwrite the current classifier parameters with the $ \theta^{t+1}_C$ .
\end{enumerate}
\textbf{Central Server}
\begin{enumerate} 
\item Accept all agent-uploaded advanced features, as well as tags corresponding to the source domain.
\item Forward propagation, calculation Eq.(\ref{con:eqall}).
\item Backpropagation, update the central server parameters of classifier and discriminator  according to Eq.(\ref{con:eqcc}) and Eq.(\ref{con:eqd}).
\item Back propagation, the central server calculates$\frac{\partial \mathcal{L}}{\partial f_S}$, $\frac{\partial \mathcal{L}}{\partial f_T}$.
\item Pass $(\frac{\partial \mathcal{L}}{\partial f_S}, \theta^{t+1}_C)$ and $(\frac{\partial \mathcal{L}}{\partial f_T},  \theta^{t+1}_C)$to the source domain and the target domain  respectively.
\end{enumerate}
\textbf{End For}
\label{con:alg_tra} }
\end{algorithm}

Through the above scheme, a fault diagnoser consisting of $F_t$ and $C$ is obtained, which is deployed at the target domain for online inference.

\section{Evaluations}
In this section, we conduct extensive experiments using real equipment fault data under different vertical FTL scenarios for performance evaluation.
Experimental methodology is first introduced, then results and analysis are presented.

\subsection{Experimental Methodology}
We validated both the \textit{effectiveness} (diagnosis accuracy) and \textit{generality} (applicability to different levels of source-target domain heterogeneity) of FedLED by comparing it with SOTA approaches under different settings.

\subsubsection{Datasets}
Our experiments used two public datasets containing different monitoring signals and fault labels of three different pieces of real equipment (two bears and a gear): \textit{i.e.}, CWRU\cite{CWRU} and Gearbox\cite{shao2018highly}.

\textbf{CWRU} is a widely adopted fault diagnosis benchmark containing three vibration signals(drive-side acceleration data DE, fan-side acceleration data FE, and the reference acceleration data BA) of an SKF6205 bear of 1067 samples. 
The vibration signal can be acquired by the accelerometer close to the motor-driven end with the 12-kHz sampling frequency. The faults with a single point are introduced to test bearings by electric discharge machining (EDM), resulting in damages of three severity with diameters of 0.007, 0.014, and 0.021 in, respectively. Depending on the location of the faults, there are three types of bearing fault, namely inner-race fault (IF), outer-race fault (OF) and ball fault (BF). Moreover, the bearing of normal condition (NC) is also tested.
For heterogeneity, the source and target domains respectively comprised two out of the three signals (features) that were not fully overlapped.
Only the source domain possessed fault labels (nine types of bear faults).
As shown in Table \ref{con:tabtask}, six different fault diagnosis tasks were selected.

\textbf{Gearbox} contains eight monitoring signals (with a 12-kHz sampling rate) of a DDS bear and gear.
The DDS consists of a brake, a planetary gearbox, a parallel gearbox, and a motor. Additionally, two three-axis (x, y, and z) acceleration sensors collect six channels of vibration signals, which are mounted on the parallel gearbox and the planetary gearbox, respectively. A torque sensor is installed between the motor and the planetary gearbox to measure load. And there are eight signal characteristics in each data file, which represent: motor vibration signal, vibration signal of planetary gearbox in three directions of x, y, and z, motor torque data and parallel gearbox in three directions of x, y, and z. vibration in one direction. According to the health status of each mechanical equipment, a total of 5115 samples were prepared. Each sample has a number of different features, depending on the task type, with 1024 data points per feature.
For heterogeneity, the source and target domains respectively comprised four/five out of the eight signals (features) that were not fully overlapped.
Only the source domain possessed fault labels (five types of faults for bear and gear).
As shown in Table \ref{con:tabger_task}, for the bear and gear, six different fault diagnosis tasks were respectively selected.

Learning samples were extracted from all monitoring signals above using the non-overlapping sliding window method \cite{dugand2019molecular}, and further divided as training and testing sets with a 7:3 ratio.
Detailed operations are provided in Supplemental Material S.3.1).

\subsubsection{Comparatives}
We compared the performance of FedLED with the following approaches.

\begin{enumerate}
\item
\textbf{Baseline}: Training the model on the source domain, and directly applying the trained model to the target domain.
\item
\textbf{SFL-multi} \cite{liu2020secure}: The only FTL-based equipment fault diagnosis method currently available.
\item
\textbf{Discrepancy-based methods}: SOTA unsupervised transfer learning methods based on different distance metrics, including CORAL \cite{sun2016deep} using covariance, MK-MMD \cite{long2018transferable} using MMD, and JAN \cite{long2017deep} using JMMD.
\item
\textbf{Adversarial-based methods}: SOTA unsupervised adversarial-based transfer learning methods, including DANN \cite{ganin2016domain} and CDA+E \cite{long2018conditional}.
\item
\textbf{Ablation study methods}: Abl Exp 1 and 2 only retained the joint domain alignment and joint domain adversarial adaptation, respectively.
\end{enumerate}

\subsubsection{Implementations}
FedLED and all comparatives were implemented using PyTorch V1.3.1, and all evaluations were conducted on a Tesla V100 GPU.
For parameter settings, the training batch size and iteration number were 64 and 100, respectively.
We use a learning rate of $lr=0.01\/(1+10 \times p)^{0.75}$, where $p \in (0,1]$ is the dynamic decaying rate.

\subsubsection{Evaluation Metrics}
We used the accuracy on the target domain testing set as the evaluation metric:
\begin{equation}
    Accuracy = \frac{n_{correct}}{n_{test}} \times 100\%,
    \label{con:eqind}
\end{equation}
where, $n_{test}$ represents all testing samples, and $n_{correct}$ is the number of all correctly diagnosed samples.
To reduce the randomness and singularity, we recorded the average accuracy of 10 repeated experiments as the final results.

\begin{table}[t]
\center
\caption{Fault Diagnosis Tasks based on CWRU}
\begin{tabular}{|m{1cm}<{\centering}|m{2cm}<{\centering}|m{2cm}<{\centering}|m{2cm}<{\centering}|}\hline
Task&	Source Domain Feature &	Target Domain Feature &	Overlapping Feature \\ \hline
T1&	FE, DE&	BA, DE&	DE \\ \hline
T2&	BA, DE&	FE, DE&	DE \\ \hline
T3&	DE, FE&	BA, FE&	FE \\ \hline
T4&	BA, FE&	DE, FE&	FE \\ \hline
T5&	DE, BA&	FE, BA&	BA \\ \hline
T6&	FE, BA&	DE, BA&	BA \\ \hline
\end{tabular}
\label{con:tabtask}
\end{table}

\begin{table}[t]
\centering
\caption{Fault Diagnosis Tasks based on Gearbox}
\begin{tabular}{|m{0.6cm}<{\centering}|m{2.7cm}<{\centering}|m{2.7cm}<{\centering}|m{1.3cm}<{\centering}|}\hline
Task&	Source Domain Feature&	Target Domain Feature&	Overlapping Feature \\ \hline
T1 & MV, PL\_x, PL\_y, PL\_z & MV, PA\_x, PA\_y, PA\_z & MV  \\ \hline
T2 & MV, PA\_x, PA\_y, PA\_z & MV, PL\_x, PL\_y, PL\_z  & MV \\ \hline
T3&	MT, PL\_x, PL\_y, PL\_z & MT, PA\_x, PA\_y, PA\_z & MT \\ \hline
T4&	MT, PA\_x, PA\_y, PA\_z & MT, PL\_x, PL\_y, PL\_z  & MT  \\ \hline
T5&	MV, MT, PL\_x, PL\_y, PL\_z & MV,MT, PA\_x, PA\_y, PA\_z & MV, MT \\ \hline
T6&	MV, MT, PA\_x, PA\_y, PA\_z & MV, MT, PL\_x, PL\_y, PL\_z  & MV, MT \\ \hline
\end{tabular}
\label{con:tabger_task}
\end{table}

\subsection{Fault Diagnosis Accuracy}
To verify the effectiveness of our method, we conducted experiments of FedLED and comparatives on CWRU and Gearbox datasets following the aforementioned tasks.
Considering that overlapping samples are mandatory to SFL-multi, we separately set a 10\% sample overlapping ratio for it, while FedLED and other comparatives were set with non-overlapping sample spaces.
The fault diagnosis accuracy of all methods is demonstrated in Figs.\ref{fig:cwru}$\sim$\ref{fig:geargear}, and detailed results are provided in Supplemental Material S.3.2).
It is obvious that FedLED achieves the highest average diagnosis accuracy (\textit{i.e.}, 77.52\%, 95.51\%, 98.47\%) on all datasets.
The performance of FedLED and comparatives are further analyzed as follows.

According to Figs.\ref{fig:cwru}$\sim$\ref{fig:geargear}, DANN performs badly on all tasks, which may be caused by the lack of initialization.
Generally, adversarial-based methods increase the domain adaptation ability of $D$ by reducing its discrimination ability.
Since DANN lacks initialization, its $D$ has a much stronger discrimination ability, severely restricting the domain adaptation ability.
SFL-multi performs stably on all tasks due to the small number of overlapping samples that avoid overfitting.

\begin{figure}[!t]
    \centering
    \includegraphics[width=3.3in]{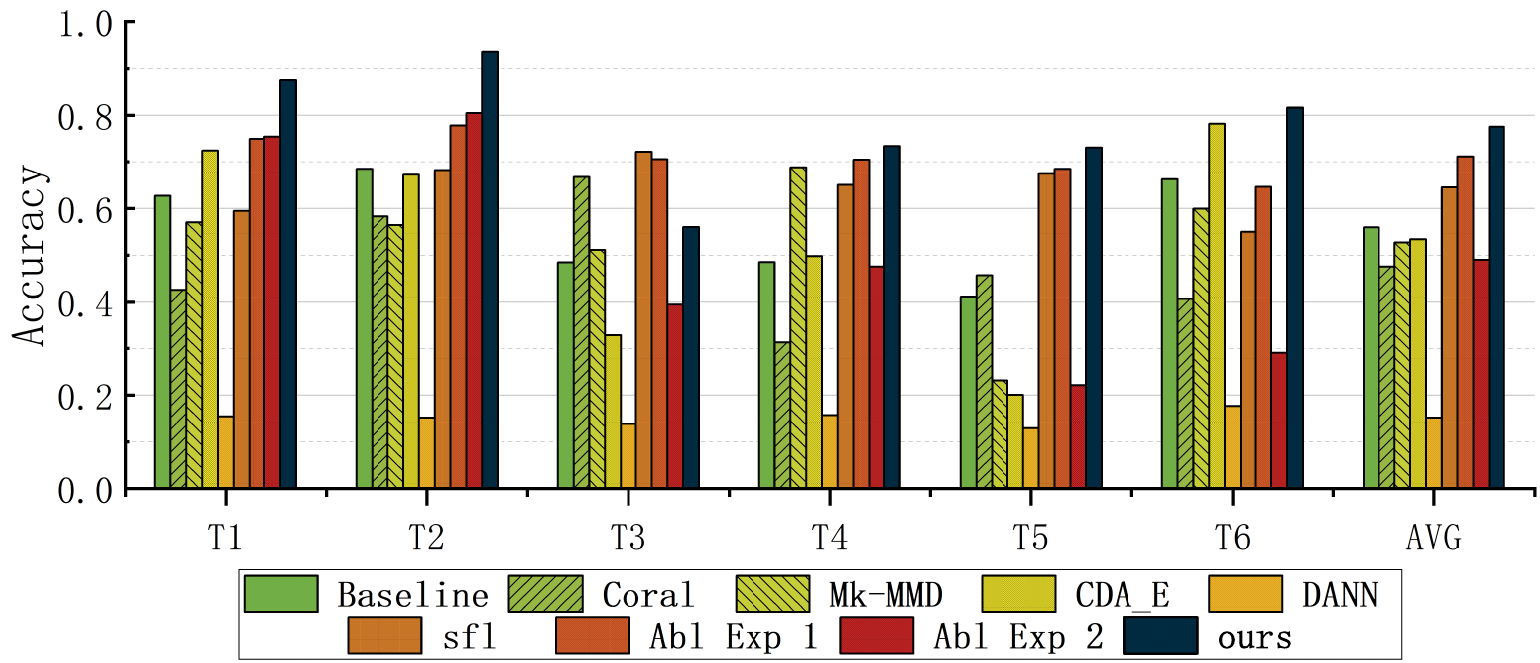}
    \caption{Fault Diagnosis Accuracy on CWRU}
    \label{fig:cwru}
\end{figure}

\begin{figure}[!t]
    \centering
    \includegraphics[width=3.3in]{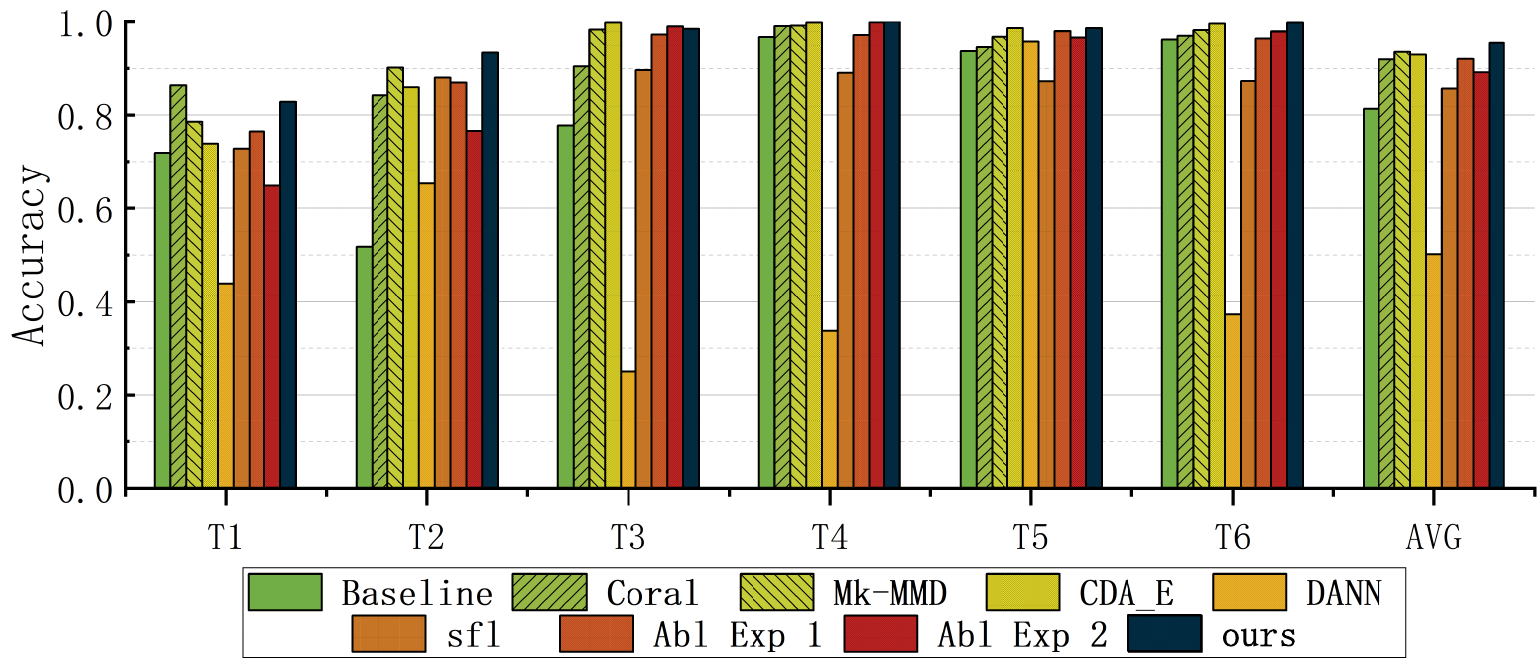}
    \caption{Fault Diagnosis Accuracy on Gearbox-bear}
    \label{fig:gearbear}
\end{figure}

\begin{figure}[!t]
    \centering
    \includegraphics[width=3.3in]{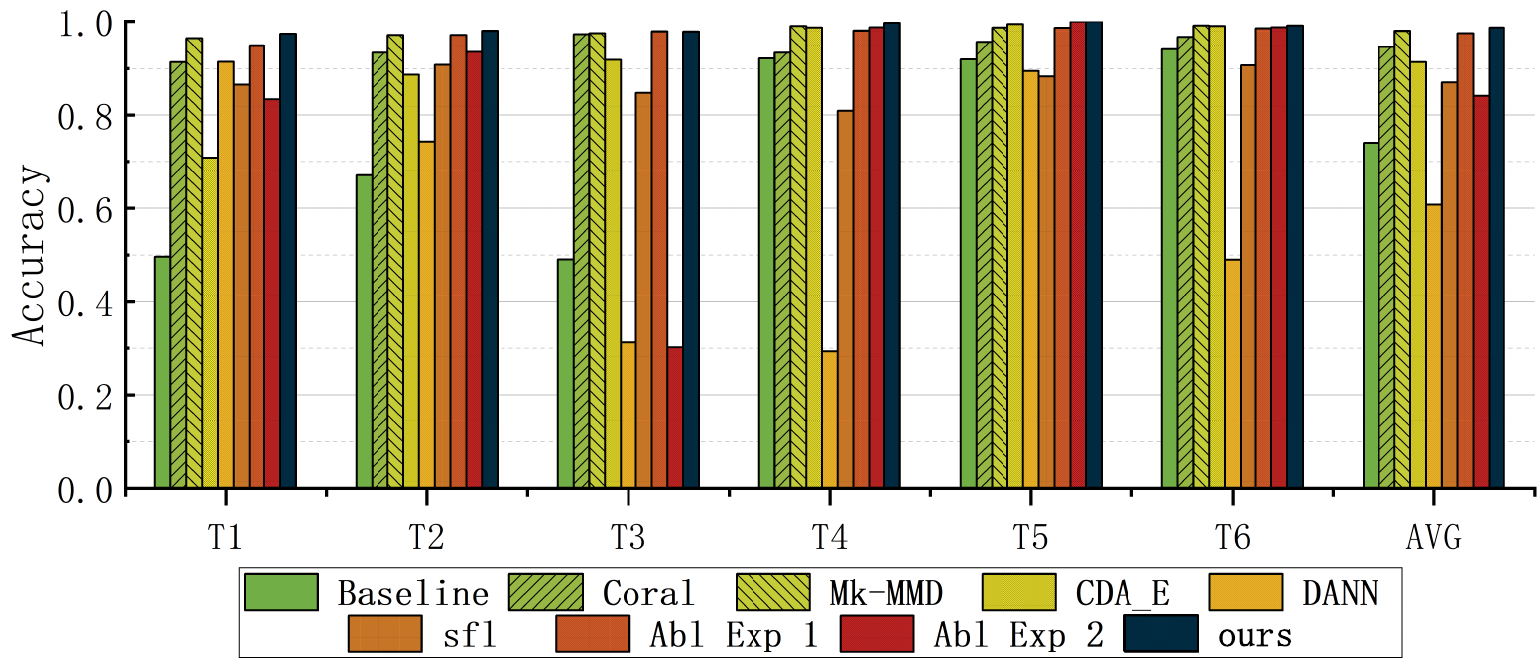}
    \caption{Fault Diagnosis Accuracy on Gearbox-gear}
    \label{fig:geargear}
\end{figure}

According to Fig.\ref{fig:cwru}, Baseline performs stably on various CWRU tasks with an average accuracy of 50.92\%, revealing the relatively low similarity between $\mathbb{R}^{N_{S}}$ and $\mathbb{R}^{N_{T}}$. 
According to Fig.\ref{fig:gearbear} and Fig.\ref{fig:geargear}, Baseline performs well on Gearbox T4$\sim$T6 while weak on Gearbox T1$\sim$T3, indicating that $dist(\mathbb{R}^{N_{S}}, \mathbb{R}^{N_{T}})$ is relatively small for T4$\sim$T6 but large for T1$\sim$T3.
Particularly, when $dist(\mathbb{R}^{N_{S}}, \mathbb{R}^{N_{T}})$ is small (\emph{e.g.} Gearbox T4$\sim$T6), adversarial-based methods (including Abl Exp 1) outperform discrepancy-based methods (including Abl Exp 2).
In the case with intense feature heterogeneity, \emph{e.g.} all CWRU tasks and Gearbox T1$\sim$T3, the performance of adversarial-based methods is weaker than discrepancy-based methods, due to the strong discrimination ability restricting the domain adaptation ability of $D$.
Considering such results, the prominent performance of FedLED clearly indicates that our introduction of joint domain alignment manages to effectively eliminate the impact of intense feature heterogeneity on general adversarial-based methods.

\subsection{Generality under Different Levels of Domain Heterogeneity}
\begin{figure}[t]
    \centering
    \includegraphics[width=2.8in]{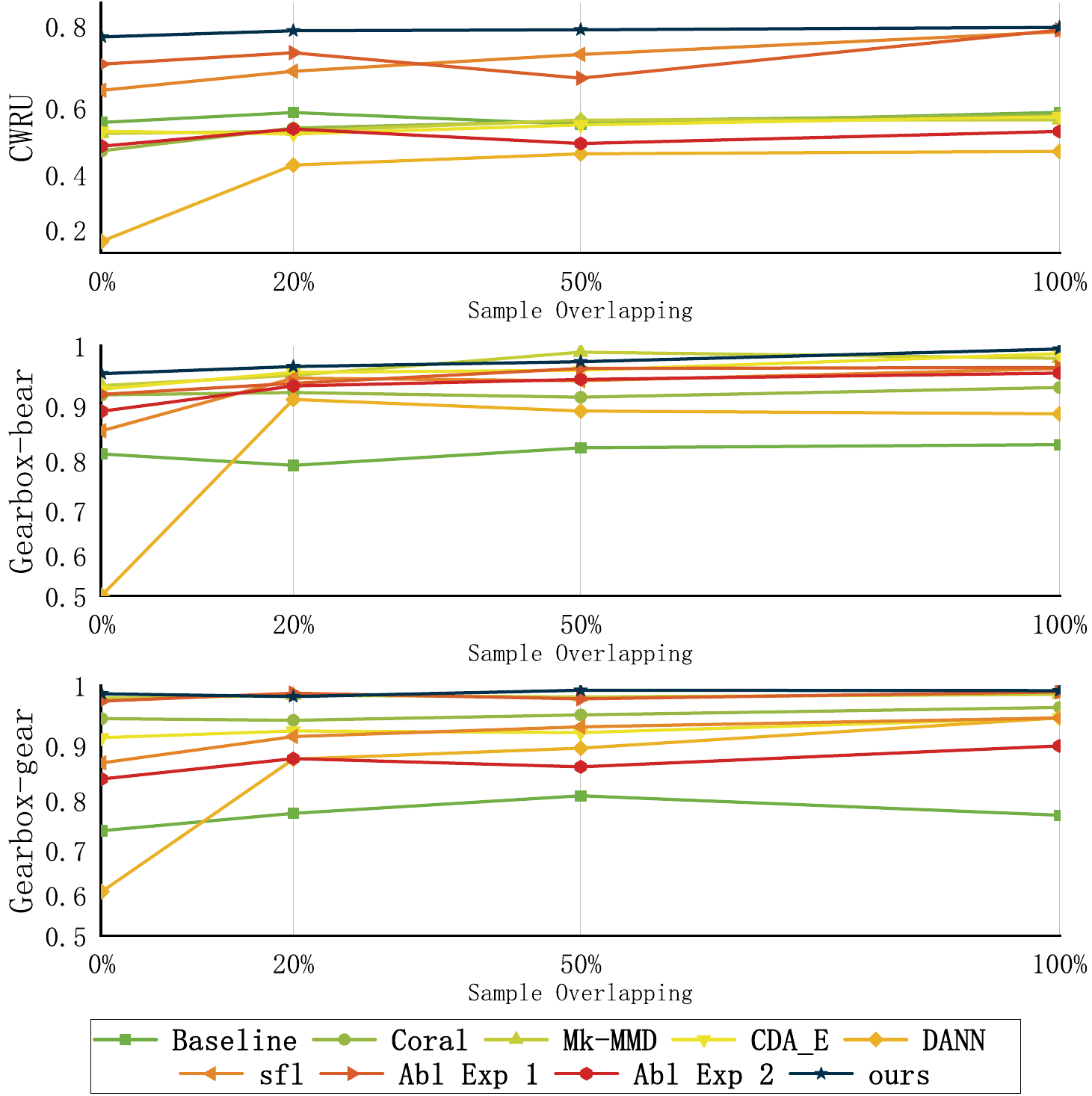}
    \caption{Fault Diagnosis Accuracy with Different Sample Overlapping Ratios}
    \label{fig:sample}
\end{figure}

To verify the generality of our method, we changed the ratios of sample and feature overlapping in the vertical FTL scenario. 
Fault diagnosis accuracy of all methods is demonstrated in Fig.\ref{fig:sample} and Fig.\ref{fig:feature_overlapping}, respectively.
Detailed results are provided in Tables S4$\sim$S7 in Supplemental Material S.3.2).

\begin{figure}[t]
    \centering
    \includegraphics[width=3.5in]{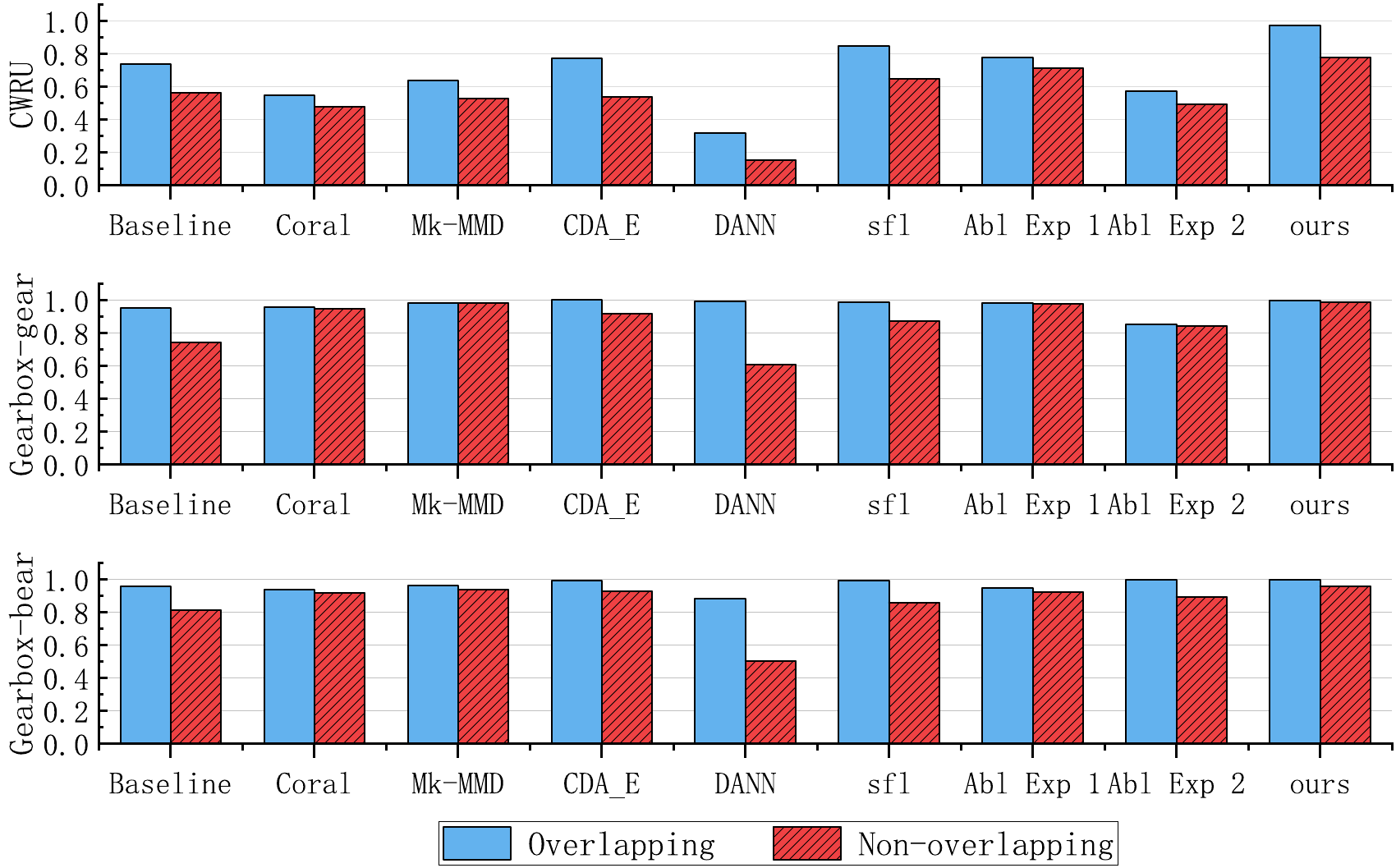}
    \caption{Fault Diagnosis Accuracy with/without Overlapping Features}
    \label{fig:feature_overlapping}
\end{figure}

\subsubsection{Impact of Sample Overlapping Ratio}
To study the impact of sample space differences, we first set the sample overlapping ratio between the source and target domains as 0\%, 20\%, 50\%, and 100\%, respectively.

According to Fig.\ref{fig:sample}, FedLED achieves the optimal performance under all sample overlapping ratios (\textit{i.e.}, 90.55\%, 91.26\%, 91.92\%, 92.76\%), and the accuracy improves slightly as the ratio increases. 
As the sample overlapping ratio increases, the accuracy of SFL-multi is significantly improved.
This is because SFL-multi needs to train a transferable model on overlapping samples, and its performance is positively related to the number of overlapping samples. 
Different from SFL-multi, the performance of other comparatives requiring no sample overlapping is not obviously enhanced.

\subsubsection{Impact of Feature Overlapping Ratio}
To study the impact of feature space differences, we conducted two sets of experiments with 0\% and 100\% feature overlapping ratios between the source and target domains, respectively.

According to Fig.\ref{fig:feature_overlapping}, FedLED achieves the highest average diagnosis accuracy (\textit{i.e.}, 98.83\%, 90.50\%) with/without feature overlapping, indicating that it manages to map different feature spaces to a latent common space.
Differently, the performance of all comparatives is significantly degraded without overlapping features, clearly demonstrating that the feature space heterogeneity of source and target domains severely restricts the usability of existing fault diagnosis approaches. 

\subsection{Result Analysis }
\begin{figure}[t]
    \centering
    \includegraphics[width=2.5in]{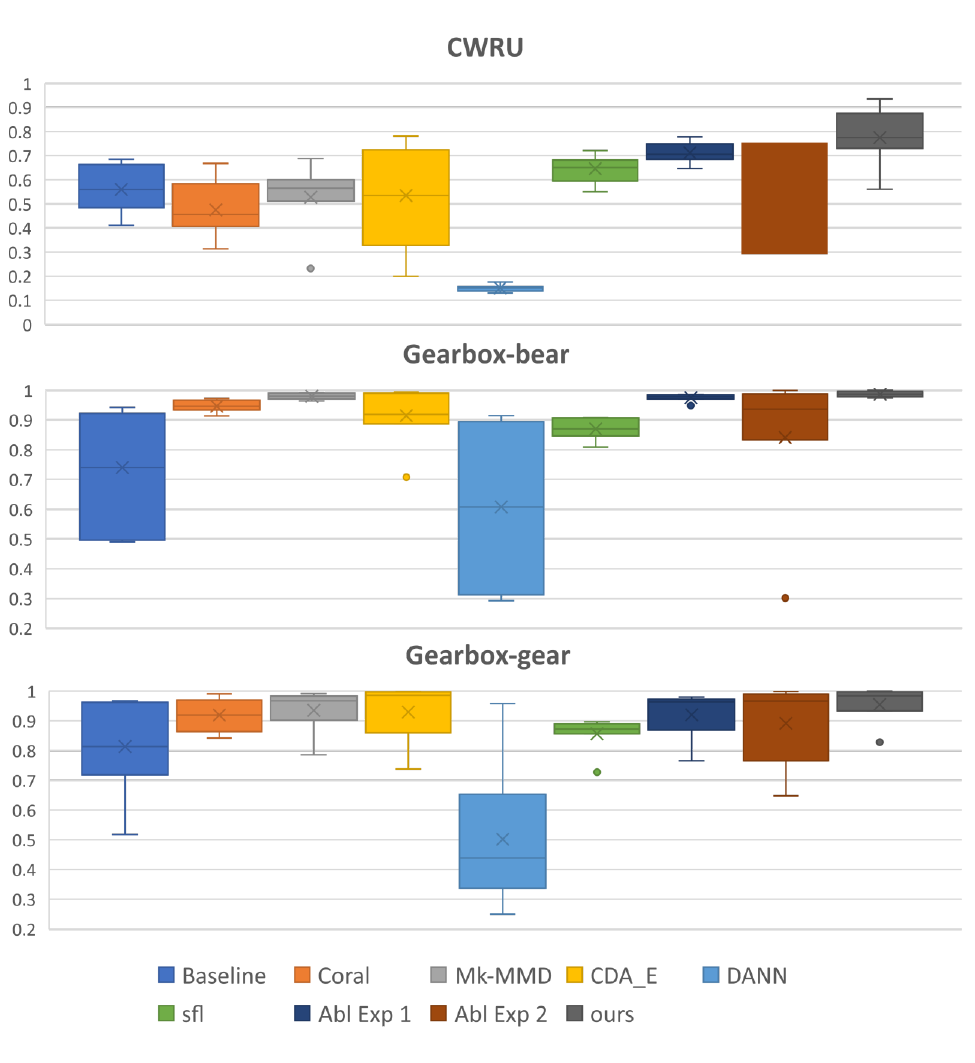}
    \caption{Box-Plot of Different Methods}
    \label{fig:xiangxing}
\end{figure}

In order to highlight the  statistical significance of our method, we performed future analysis of above experimental results on the stability and complexity. Fig.\ref{fig:xiangxing} shows the box-plot of different methods under all three datasets, reflecting the statistical characteristics of their accuracy. It can be found that our method has relatively good stability and meets the needs of practical applications.

\begin{figure}[t]
    \centering
    \subfloat[\footnotesize Trend Chart of Loss]{
    \includegraphics[width=1.7in]{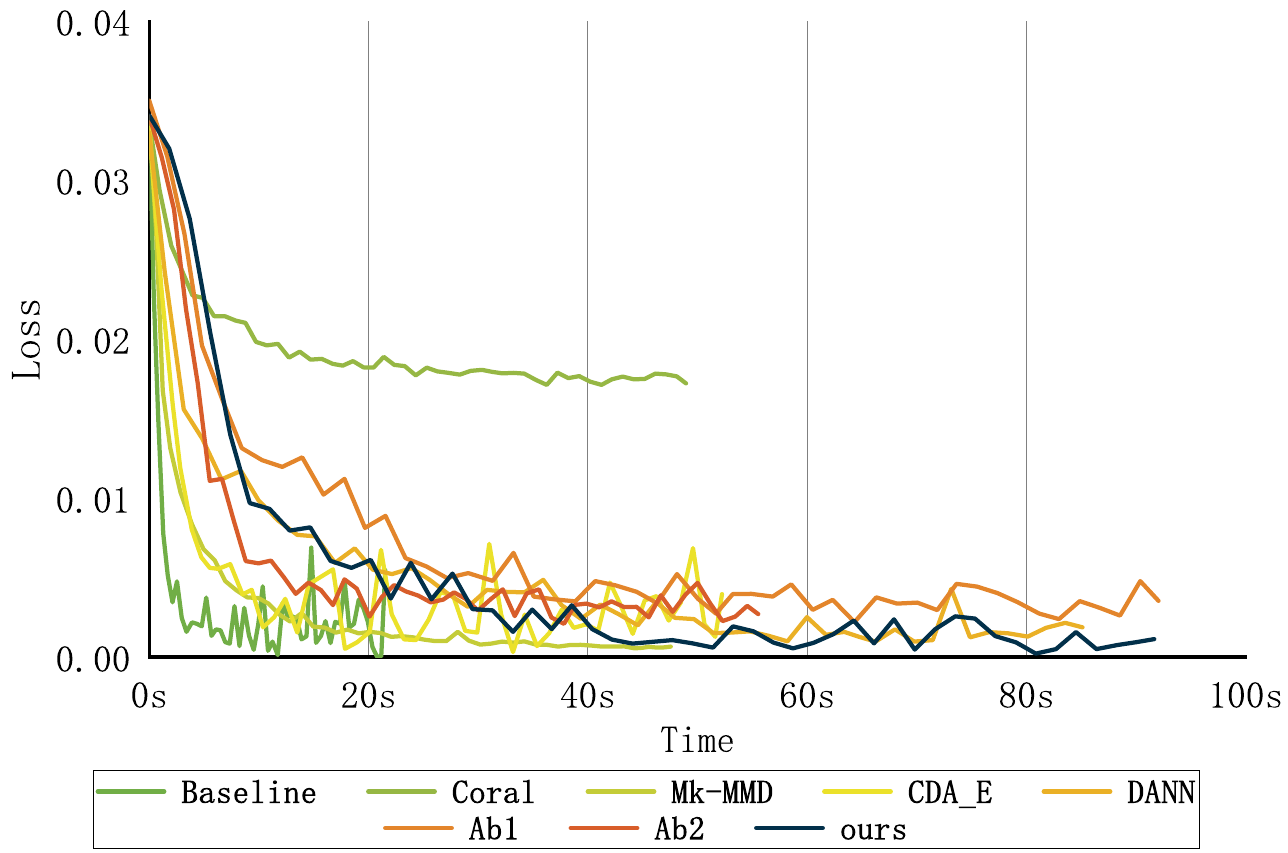}}
    \subfloat[\footnotesize Trend Chart of Accuracy]{
    \includegraphics[width=1.7in]{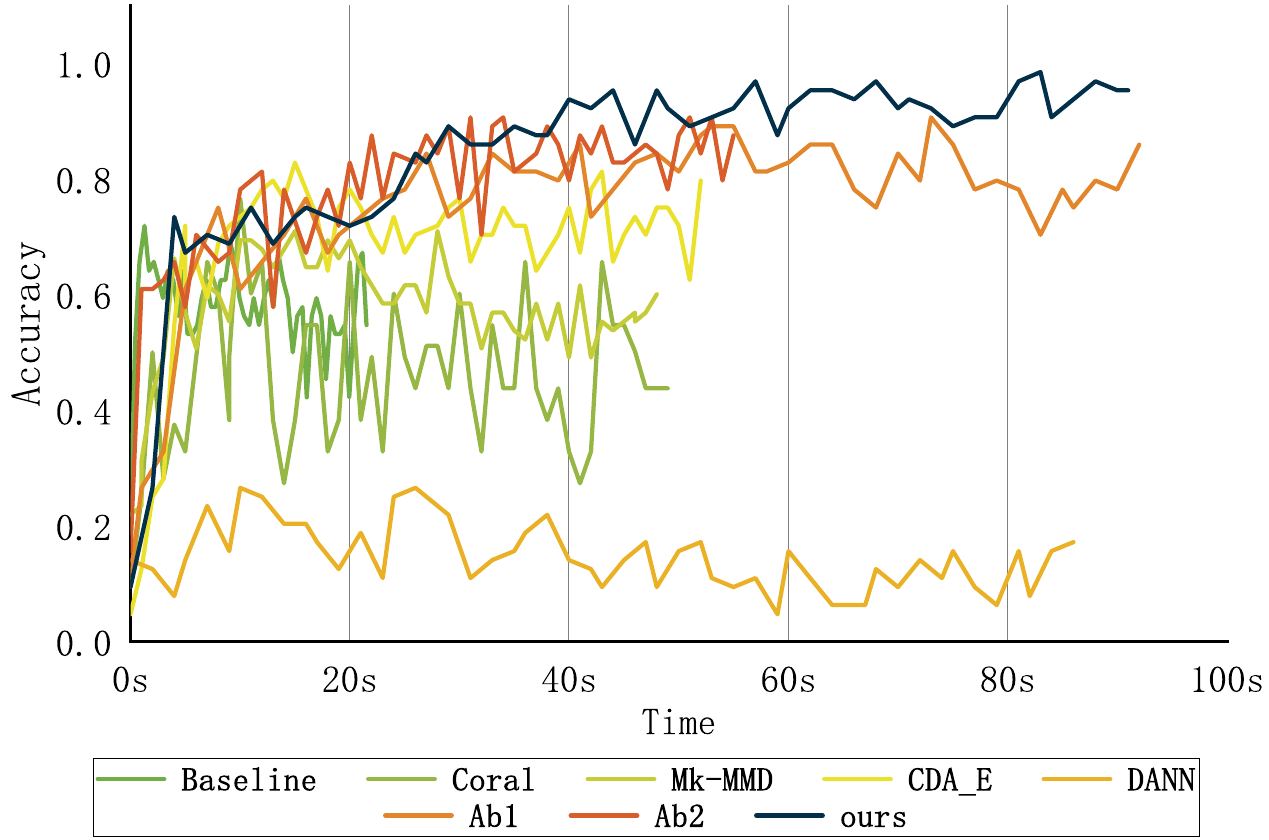}}
    \caption{Trend Charts over Time under CWRU}
    \label{fig:timecon}
\end{figure}

We counter the average time of our method and  comparatives running for 100 epochs under the CWRU and Gearbox datasets, which are recorded in Table \ref{con:time}.  
The original SFL method is only suitable for two-classification problems. We have extended it to multi-classification problems through multiple classifiers, which also results in a much more time consumption than other methods.
Fig.\ref{fig:timecon} (a) and (b) are  the trend charts of loss and accuracy over time under CWRU. It can be found that the overall convergence speed of our method is slightly slower than other comparisons due to the longer training time of each epoch. However, because of the improvement in accuracy, the time it takes for our method to achieve the same acceptable accuracy (70\%) is actually similar compared with other comparatives.

Network complexity includes space complexity and time complexity. The complexity of our FedLED network is on the same order of magnitude with other comparatives except SFL
Our method and other comparatives except SFL are based on the same deep transfer learning backbone, with an  additional 3-layer domain discriminator whose space complexity is negligible compared to the backbone, so the space complexity of our method is similar to other comparatives.
As for time complexity, FedLED consists of the vertical federated joint domain adversarial adaptation and the joint domain alignment, their time complexity corresponds to adversarial-based and discrepancy-based methods respectively. Therefore, the time complexity of our method is the sum of the two parts’ time complexity, which means it is on the same order of magnitude with the time complexities of other comparatives. 
To sum up, although our method is slightly more time-consuming than other comparatives, compared with the improvement in accuracy, such trade-off is acceptable and worthwhile.

\begin{table}[t]
\center
\caption{Average Consuming Time of Different Methods for 100 Epoch Training (Sec)}
\setlength{\tabcolsep}{4.5mm}{
\begin{tabular}{c|c|c}\hline
Methods&	CWRU&	Gearbox \\ \hline
Baseline&	4&	22 \\ \hline
CORAL&	7&	49 \\ \hline
Mk-MMD&	10&	54\\ \hline
CDA+E&	9&	49 \\ \hline
DANN&	14&	82 \\ \hline
SFL-multi&	168&	884 \\ \hline
Abl Exp 1&	17&	92 \\ \hline
Abl Exp 2&	8&	52 \\ \hline
Ours&	18&	94 \\ \hline
\end{tabular}}
\label{con:time}
\end{table}

\section{Conclusion}
In this paper, we present FedLED, the first unsupervised vertical FTL method facilitating a wide range of industrial agents to conduct label-free equipment fault diagnosis.
It enables transferring a fault diagnosis model from a labeled source domain to a highly heterogeneous target domain with zero fault label while preserving the data privacy of both domains.
Extensive experiments using real equipment monitoring data clearly demonstrate that FedLED manages to achieve obvious advantages in terms of both diagnosis accuracy (up to 4.13$\times$ higher) and generality by exploiting knowledge from the unlabeled target domain, different from SOTA approaches intensely depending on source domain knowledge.
We expect FedLED to inspire more insights on label-free fault diagnosis enhanced by systematic target domain knowledge extraction, \textit{e.g.}, contrastive learning.

\bibliographystyle{IEEEtran}
\bibliography{cas-refs}

\end{document}